\documentclass[table, 11pt]{article}
\usepackage[table, svgnames, dvipsnames]{xcolor} % Gộp xcolor
\usepackage{tikz}
\usepackage[justification=centering]{caption}

\usepackage[colorlinks=true, citecolor=blue, linkcolor=blue, urlcolor=blue]{hyperref}
\usepackage{xurl}
\usepackage[final]{acl}
\usepackage{amsfonts} % For \mathbb{R}
\usepackage{booktabs} % For better table rules
\usepackage{multirow} % If needed for table structure, though not strictly for this one
\usepackage{graphicx} % ở phần preamble nếu chưa có
\usepackage{amsfonts} % For \mathbb{R}
\usepackage{booktabs} % For professional tables
\usepackage{multirow} % For multi-row cells in tables
\usepackage{graphicx} % Required for includegraphics (though not used in text here)
\usepackage{xcolor} % For coloring text
\usepackage{bold-extra} % For bold text if needed, or just use \textbf
\usepackage{times}
\usepackage{booktabs}
\usepackage{multirow}
\usepackage{graphicx}
\usepackage{array}
\usepackage{latexsym}
\usepackage{enumitem}
\usepackage[T1]{fontenc}
\usepackage[utf8]{inputenc}
\usepackage{microtype}
\usepackage{inconsolata}
\usepackage{graphicx}
\usepackage{amsmath}
\usepackage{amsfonts}
\usepackage{bm}
\usepackage{multirow}
\usepackage{booktabs}
\usepackage{xspace}
\usepackage{adjustbox}
\usepackage{float}
\usepackage{natbib} % Thêm natbib để hỗ trợ \citep, \citet

\usepackage{colortbl}
\usepackage{subfigure}

\usepackage{colortbl}
\usepackage{subfigure}

\definecolor{lightgray}{gray}{0.80}
\definecolor{darkgreen}{rgb}{0.0, 0.5, 0.0}

\newcolumntype{g}{>{\columncolor{lightgray}}c}
\newcommand*{\method}{\textsc{MIPIC}\@\xspace}

\title{\method: Matryoshka Representation Learning via Self-Distilled Intra-Relational and Progressive Information Chaining}

\author{
    \textbf{Phung Gia Huy\textsuperscript{1}\footnotemark[1]},
  \textbf{Hai An Vu\textsuperscript{1}\footnotemark[1]},   
   \textbf{Minh-Phuc Truong\textsuperscript{1}\footnotemark[1]},
    \textbf{Thang Duc Tran\textsuperscript{1}},\\
    \textbf{Linh Ngo Van\textsuperscript{1\dag}},
  \textbf{Thanh Nguyen Hong\textsuperscript{2}},
  \textbf{Trung Le\textsuperscript{3}}
  \bigskip \\
\textsuperscript{1}Hanoi University of Science and Technology, \\
\textsuperscript{2}University of Oregon, 
\textsuperscript{3}Monash University,
}

\begin{document}

\maketitle
\renewcommand{\thefootnote}{\fnsymbol{footnote}}
\footnotetext[1]{Equal contribution}
\footnotetext[2]{Corresponding author: \href{mailto:email@domain}{ linhnv@soict.hust.edu.vn}}
\renewcommand*{\thefootnote}{\arabic{footnote}}
\begin{abstract}
    Representation learning is fundamental to NLP, but building embeddings that work well at different computational budgets is challenging. Matryoshka Representation Learning (MRL) offers a flexible inference paradigm through nested embeddings; however, learning such structures requires explicit coordination of how information is arranged across embedding dimensionality and model depth. In this work, we propose MIPIC (Matryoshka Representation Learning via Self-Distilled Intra-Relational Alignment and Progressive Information Chaining), a unified training framework designed to produce structurally coherent and semantically compact Matryoshka representations. MIPIC promotes cross-dimensional structural consistency through Self-Distilled Intra-Relational Alignment (SIA), which aligns token-level geometric and attention-driven relations between full and truncated representations using top-k CKA self-distillation. Complementarily, it enables depth-wise semantic consolidation via Progressive Information Chaining (PIC), a scaffolded alignment strategy that incrementally transfers mature task semantics from deeper layers into earlier layers. Extensive experiments on STS, NLI, and classification benchmarks (spanning models from TinyBERT to BGEM3, Qwen3) demonstrate that MIPIC yields Matryoshka representations that are highly competitive across all capacities, with significant performance advantages observed under extreme low-dimensional.
\end{abstract}
\section{Introduction}\label{sec:intro}
Learned dense representations are the cornerstone of modern NLP \citep{mikolov2013distributedrepresentationswordsphrases, pennington-etal-2014-glove, devlin2019bertpretrainingdeepbidirectional}, yet their high computational demands often limit deployment \citep{wang2020minilmdeepselfattentiondistillation}. Matryoshka Representation Learning (MRL) \citep{NEURIPS2022_c32319f4} addresses this by introducing nested embeddings that enable adaptive inference-time truncation, allowing a single model to satisfy diverse computational budgets without retraining. Training effective Matryoshka representations, however, presents challenges that extend beyond conventional embedding learning.
Unlike standard representations, MRL requires semantic information to be explicitly organized such that meaningful structure is preserved under progressive dimensional truncation.
Existing MRL formulations
\citep{NEURIPS2022_c32319f4, li2025ese}
primarily enforce nested usability through independent supervision at different truncation levels or sentence-level alignment objectives.
While such strategies encourage prefix usability, they leave open the broader question of how semantic information should be internally arranged to support robust Matryoshka structures.
In particular, learning low-dimensional prefix embeddings that remain semantically expressive requires the model to carefully organize semantic features across both embedding dimensions and network layers, involving a non-trivial restructuring of how information is represented and propagated.

First, effective MRL can be viewed through the lens of \textbf{cross-dimensional structural alignment}, where low-dimensional prefixes are encouraged to reflect not only sentence-level outputs but also the internal relational structure (like token-level geometric relationships, hidden states) of the full-dimensional representations. From this perspective, aligning relational patterns across different dimension spaces provides a natural way to support stable semantic behavior in compact prefixes. Second, MRL may also be understood from the perspective of involving \textbf{depth-wise semantic consolidation}, in which task-relevant information is gradually condensed from deeper layers, where representations are typically more expressive, into earlier layers. This process is inherently progressive: rather than emerging at a single truncation point, core semantic features can be refined and transferred as information flows through the network. Viewing Matryoshka training through this lens highlights the benefits of intermediate guidance that supports a smooth, coarse-to-fine refinement of representations across depth.

Motivated by these considerations, we propose \textbf{MIPIC} (\textbf{M}atryoshka Representation Learning via Self-Distilled \textbf{I}ntra-Relational Alignment and \textbf{P}rogressive \textbf{I}nformation \textbf{C}haining), a unified training framework that explicitly organizes information across both embedding dimensionality and model depth, enabling the learning of structurally coherent and semantically compact Matryoshka representations. To encourage cross-dimensional structural consistency, MIPIC employs Self-Distilled Intra-Relational Alignment (SIA). Rather than aligning representations solely at the sentence level like prior MRL approaches \citep{NEURIPS2022_c32319f4, li2025ese}, SIA focuses on token-level intra-relations derived from geometric similarity and attention-based importance. Using Centered Kernel Alignment (CKA) \citep{kornblith2019similarityneuralnetworkrepresentations} with a hard top-$k$ selection strategy, SIA selectively aligns the most salient relational patterns between full and truncated representations. This targeted self-distillation goes beyond simple alignment; it effectively reorganizes the internal feature hierarchy, forcing critical semantic structures to be prioritized and concentrated within the low-dimensional prefix dimensions. Complementing cross dimensional alignment, MIPIC introduces Progressive Information Chaining (PIC) to guide the flow of semantic information across the network depth. Unlike prior methods that restrict task supervision solely to the final representation, which can be unstable when compressing rich knowledge into narrow dimensions, PIC builds a scaffolded alignment pipeline in which each step acts as a semantic bridge. This design is motivated by empirical findings that lower Transformer layers predominantly encode low level linguistic features such as syntax and basic semantics, while higher layers increasingly capture task specific and discriminative signals \citep{jawahar-etal-2019-bert, liu-etal-2019-linguistic}. By propagating task aware cues from upper layers into earlier ones, PIC performs a controlled early intervention that gently biases the formation of low level representations toward task relevant subspaces, ensuring that useful discriminative structure is available from the beginning of the network processing. Importantly, by aligning only the most essential components of each layer representation, which have been reorganized by SIA to concentrate the most critical task relevant knowledge into a compact subspace, PIC allows the remaining dimensions to continue modeling low level and general linguistic features. This selective supervision preserves the model natural representational flexibility, avoids over regularization, and enables a stable and progressive refinement of task relevant information across depth. Together, SIA and PIC provide a principled training framework for MRL that explicitly organizes semantic information across both dimensionality and depth.

Our main contributions are:
\begin{itemize}
    \item We propose \textbf{MIPIC}, a unified framework that addresses structural and semantic incoherence in MRL by explicitly coordinating cross-dimensional alignment and depth-wise information consolidation.
    \item We introduce two complementary mechanisms: Self-Distilled Intra-Relational Alignment (SIA), which utilizes top-k CKA self distillation to enforce cross-dimensional topological consistency via self-distillation, and Progressive Information Chaining (PIC), a scaffolded learning strategy that enables lower-dimensional representations to acquire task-relevant information early in training through progressive semantic guidance across network depth.
    \item We present extensive experiments on STS, NLI, and text classification benchmarks showing that MIPIC significantly improves representation efficiency across diverse backbones, from TinyBERT-6L and BERT-base to large-scale models such as Qwen3 embedding 0.6B and BGE-M3. MIPIC remains competitive at full capacity while consistently outperforming state-of-the-art baselines under severe truncation, particularly in low-dimensional regimes.
\end{itemize}

\section{Related Work and Background}
\subsection{Related Works}
\paragraph{Learning Sentence Embeddings}

Sentence embeddings have evolved from static word vectors \citep{pennington-etal-2014-glove, mikolov2013distributedrepresentationswordsphrases} and contextual models \citep{peters-etal-2018-deep, devlin2019bertpretrainingdeepbidirectional}, which initially lacked optimized sentence-level semantics. Sentence-BERT \citep{reimers2019sentencebertsentenceembeddingsusing} addressed this via siamese fine-tuning, dramatically improving retrieval efficiency. Subsequent work introduced contrastive learning \citep{gao2022simcsesimplecontrastivelearning, nishikawa-etal-2022-ease} to enhance robustness, while recent advances leverage Large Language Models to further refine representation quality \citep{he2025refiningsentenceembeddingmodel, llm2vec, tao2025llmseffectiveembeddingmodels,li2025your}. More recently, embedding model distillation (or LLM distillation) has emerged as a promising direction to transfer knowledge from large models into efficient embedding architectures, leveraging techniques such as intra-model relational alignment, optimal transport, and layer-wise mixtures to enhance representation quality \citep{truong2025emo, an2026mol, truong2026ctpd}. These distilled embeddings have demonstrated strong effectiveness across downstream applications, including cross-lingual retrieval, retrieval-augmented generation, event detection, and continual relation extraction, topic modelling \citep{hieu2025magix, nguyen2025improving, hai2026mozila, anh2025mutual, le2025enhancing, pham2025mitigating, phat2025extra, phat2026gloctm}.

\paragraph{Matryoshka Representation Learning methods}

While sentence embeddings have advanced significantly, their fixed dimensionality remains a computational bottleneck at scale. Matryoshka Representation Learning (MRL) addresses this by learning nested coarse-to-fine prefixes within a single embedding, enabling low-dimensional prefixes to function as standalone representations with only $O(\log d)$ supervision points and no additional forward passes \citep{NEURIPS2022_c32319f4}. This design enables substantial compression with minimal accuracy loss and faster inference. Building on MRL, Espresso Sentence Embeddings (ESE) further enhance scalability across depth and dimensionality through learn to express and learn to compress mechanisms \citep{li2025ese}. Beyond sentence embeddings, the Matryoshka principle has been applied to image generation \citep{gu2024matryoshkadiffusionmodels}, multimodal representation learning \citep{cai2024matryoshkamultimodalmodels}, and multimodal LLMs \citep{hu2024matryoshkaquerytransformerlarge}.

\subsection{Background}

\subsubsection{Matryoshka Representation Learning}

Let an encoder \(f_\theta\) map an input sentence \(x\) to a high-dimensional embedding \(z=f_\theta(x)\in\mathbb{R}^{D}\). In Matryoshka Representation Learning \citep{NEURIPS2022_c32319f4} we fix an ordered set of nested prefix dimensions
$
\mathcal{D}=\{d_1,d_2,\dots,d_n\}, 0<d_1<d_2<\cdots<d_n=D,
$
so that for each \(d_i\in\mathcal{D}\) the truncated embedding is the prefix \(z^{(d_i)}:=z_{1:d_i}\in\mathbb{R}^{d_i}\). Each prefix \(z^{(d_i)}\) is treated as an independent representation used by a (possibly shared) task head \(g_{\phi_i}:\mathbb{R}^{d_i}\to\mathcal{Y}\) and evaluated with a task loss \(\mathcal{L}_{\text{task}}^{(d_i)}(x,y)=\ell\big(g_{\phi_i}(z^{(d_i)}),y\big)\) for a target \(y\). The Matryoshka training objective jointly supervises all prefixes; a common choice is the unweighted sum

\begin{equation}
\label{MRL loss}  
\mathcal{L}_{\text{MRL}}(\theta,\{\phi_i\}) \;=\; \sum_{i=1}^{n}\; \mathbb{E}_{(x,y)\sim\mathcal{D}_{\text{train}}}\!\big[\mathcal{L}_{\text{task}}^{(d_i)}(x,y)\big]
\end{equation}
or, when desired, a weighted variant \(\sum_{i=1}^n \alpha_i\,\mathbb{E}[\mathcal{L}_{\text{task}}^{(d_i)}]\) with \(\alpha_i\ge0\). By optimizing this objective, Matryoshka models learn a single high-dimensional embedding whose low-dimensional prefixes \(z^{(d_1)},\dots,z^{(d_{n-1})}\) are immediately usable as standalone embeddings at inference time, avoiding extra forward passes.

Matryoshka Representation Learning (MRL) has recently emerged as an effective paradigm for structuring embedding spaces such that meaningful representations are preserved across multiple prefix dimensions \citep{kusupati2024matryoshkarepresentationlearning, li2025your}. Instead of optimizing solely for full-dimensional performance, MRL explicitly enforces that lower-dimensional subspaces remain semantically informative, enabling flexible trade-offs between efficiency and accuracy. This property is particularly beneficial in large-scale retrieval and resource-constrained scenarios, where smaller embeddings can significantly reduce memory footprint and computation while maintaining competitive performance. Prior works have shown that while full-dimensional performance is often comparable to standard embedding approaches, the key advantage of MRL lies in its robustness under aggressive dimensional truncation, making it a practical solution for adaptive and efficient inference.

\subsubsection{Measuring Representational Similarity: Centered Kernel Alignment (CKA)}
\label{CKA}

Canonical Correlation Analysis (CCA) \citep{6788402} has been widely used to compare learned representations by seeking linear projections that maximize correlation between two feature sets. However, CCA can be fragile in practice: it is sensitive to simple transformations of the features and can be costly to compute for very high-dimensional activations common in deep models \citep{kornblith2019similarityneuralnetworkrepresentations}.  Centered Kernel Alignment (CKA) \citep{kornblith2019similarityneuralnetworkrepresentations} provides a more robust and efficient alternative by shifting the comparison from individual feature coordinates to the \emph{pairwise similarity structure} induced by each representation. Given two representation matrices $\mathbf{X}\in\mathbb{R}^{m\times S}$ and $\mathbf{Y}\in\mathbb{R}^{m\times T}$ for the same $m$ inputs, CKA builds kernel (Gram) matrices $\mathbf{K}$ and $\mathbf{L}$ with entries $K_{ij}=k(x_i,x_j)$ and $L_{ij}=l(y_i,y_j)$, and measures their dependence via the Hilbert–Schmidt Independence Criterion (HSIC) after centering:
\begin{equation}
    \mathrm{HSIC}(\mathbf{K},\mathbf{L})=\frac{1}{(m-1)^2}\operatorname{tr}(\mathbf{K}\mathbf{H}\mathbf{L}\mathbf{H})
\end{equation}
where $\mathbf{H}=I_m-\tfrac{1}{m}\mathbf{1}\mathbf{1}^\top$ removes mean effects. CKA then normalizes HSIC to obtain:
\begin{align}
    \mathrm{CKA}(\mathbf{X},\mathbf{Y})=\frac{\mathrm{HSIC}(\mathbf{K},\mathbf{L})}{\sqrt{\mathrm{HSIC}(\mathbf{K},\mathbf{K})\,\mathrm{HSIC}(\mathbf{L},\mathbf{L})}}
\end{align}
% CKA is well-suited for Matryoshka learning due to its invariance to scaling and orthogonal transformations, which prioritizes geometric structure over raw coordinates \citep{kornblith2019similarityneuralnetworkrepresentations}. Crucially, its kernel-based formulation naturally accommodates differing dimensionalities, enabling us to enforce consistent relational geometries between low-dimensional prefixes and full embeddings.

% Two properties make CKA particularly well suited for Matryoshka representation learning. First, the normalization confers invariance to isotropic scaling and orthogonal transformations, so CKA focuses on the geometry of the representation space rather than on raw coordinate values. Second, because CKA compares similarity (kernel) matrices, it naturally handles cases where the two representations have different feature dimensionalities. This latter point is crucial for Matryoshka settings, where we compare low-dimensional prefixes to full-dimensional embeddings: CKA evaluates whether the two representations arrange the inputs in the same relational geometry, regardless of their differing lengths or bases.

\section{Methodology}
\begin{figure*}[t]
    \centering
    \includegraphics[width=1.0\linewidth]{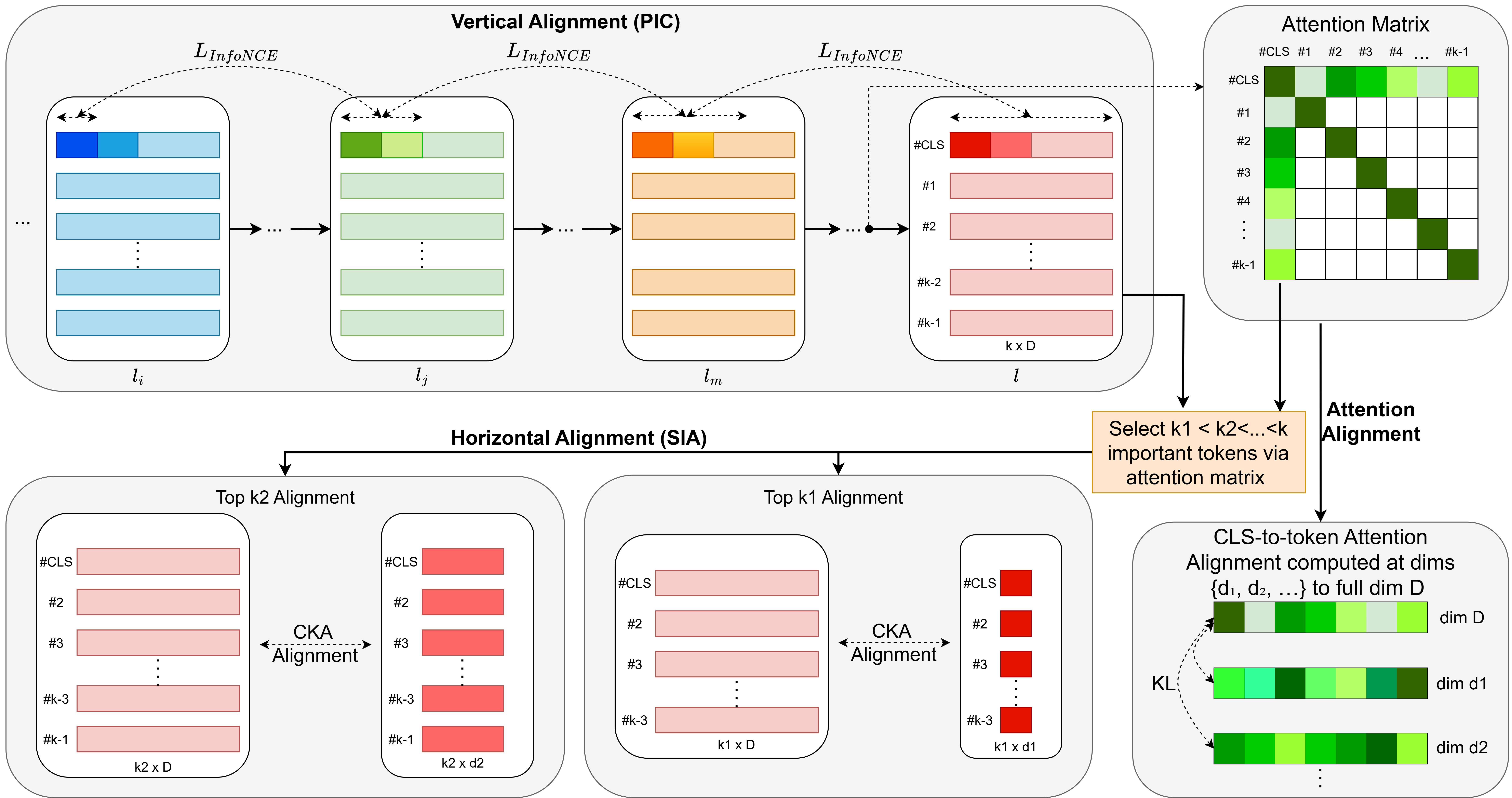}
    \caption{\textbf{Overview of MIPIC:} The framework operates via two mechanisms:\textbf{Vertical Alignment (PIC)} and \textbf{Horizontal Alignment (SIA)}}
    \label{MIPIC}
\end{figure*}

We introduce \textbf{MIPIC}, a framework designed to synchronize the learning of nested representations with the model's internal information flow. Our approach operates through two synergistic mechanisms: Self-Distilled Intra-Relational Alignment (SIA) reorganizes the internal representation hierarchy to prioritize core structural features within low-dimensional subspaces, and Progressive Information Chaining (PIC) establishes a scaffolded pipeline to condense task-specific information into these prioritized dimensions across the network's depth. Together, these components ensure that the resulting embeddings are both structurally consistent and semantically rich across all granularities. The overall architecture is shown in Figure \ref{MIPIC}.

\subsection{Preliminaries and Notation}
Consider a Transformer-based encoder with $L$ layers. Let the input sequence length be $m$, and the model hidden dimension be $D$. We define a strictly increasing sequence of nested feature sizes $\mathcal{D} = \{d_1 < d_2 < \dots < d_n = D\}$ and a corresponding sequence of layer  $\mathcal{L} = \{l_1 < l_2 < \dots < l_n\}$ at which both SIA and PIC mechanisms are applied.

\subsection{SIA: Self-Distilled Intra-Relational Alignment}
% Standard Matryoshka representation learning methods typically enforce nestedness by directly minimizing distances between sentence-level output vectors, largely neglecting the rich token-level information encoded within the model. To address this, we propose \textbf{SIA}, which treats the alignment of low-dimensional prefixes not merely as a distance-matching task, but as a \textbf{reorganization of internal information}. We formulate this process as a form of \textbf{layer-wise self-distillation}, where the full-dimensional representation at each layer serves as the teacher and its lower-dimensional prefix acts as the student, inheriting only the most essential structural signals. SIA explicitly enforces that truncated subspaces inherit the intra-relational structure of the full-dimensional representation, ensuring that core structure is prioritized and packed into the prefix dimensions. This component consists of two mechanisms: Attention Distribution Matching and Top-$k$ Hidden State Alignment via CKA.

Standard Matryoshka methods enforce nestedness by minimizing distances between sentence-level vectors, often ignoring token-level information. We propose SIA, which frames low-dimensional prefix alignment as a reorganization of internal information rather than mere distance matching. Using layer-wise self-distillation, each layer’s full-dimensional representation teaches its lower-dimensional prefix, transferring only essential structural signals. SIA ensures truncated subspaces preserve the intra-relational structure of the full representation, emphasizing core information. This is achieved via Attention Distribution Matching and Top-$k$ Hidden State Alignment with CKA.

\subsubsection{Attention Distribution Matching}
The core intuition of this module is to preserve the \textbf{relative importance ordering} of tokens within a sequence. We adopt a self-distillation strategy where the full-dimensional representation acts as a teacher to guide the lower-dimensional representation.  Let $h_{\text{CLS}} \in \mathbb{R}^{D}$ and $h_j \in \mathbb{R}^{D}$ denote the hidden states of the \texttt{[CLS]} token and the $j$-th contextual token in the full dimension $D$. The teacher's attention scores $s^{(D)}_j$ and the resulting distribution $a_D$ serve as the ground truth for alignment:
\begin{equation}
    s_j^{(D)} = \frac{ h_{\text{CLS}}\cdot h_j }{\sqrt{D}}, \quad a_D = \text{softmax}\left(\frac{s^{(D)}}{\tau}\right)
\end{equation}
For dim $d_i$, sliced tokens $h_j[1:d_i]$ are up-projected via a learnable matrix $P_i \in \mathbb{R}^{d_i \times D}$. We use $h_{\text{CLS}}$ as the query anchor since it encodes global sequence semantics, deriving attention scores $s_j^{(i)}$ that quantify each token's contribution:

\begin{equation}
    s_j^{(i)} = \frac{\\ h_{\text{CLS}} \cdot P_i ^{\top}h_j^{(i)}  }{\sqrt{D}}, \quad a_i = \text{softmax}\left(\frac{s^{(i)}}{\tau}\right)
\end{equation}
The matrix $P_i$ is trained jointly with the backbone to minimize the Kullback-Leibler (KL) divergence between the student and teacher distributions:
\begin{equation}
    \mathcal{L}_{\text{att}}^{(i)} = \sum_{d_i \in \mathcal{D}} \text{KL}(a_i \| a_D)
\end{equation}
We use the full-dimensional $h_{\text{CLS}}$ as a fixed semantic reference during training. The similarity between each token and this [CLS] vector serves as an estimate of the token importance. By aligning these similarities across dimensions, we enforce a \textbf{ranking consistency constraint}: the lower-dimensional representation is encouraged to preserve the same ordering of important tokens learned by the full-dimensional model, so that the most informative tokens remain emphasized even after compression. 

\subsubsection{Top-$k$ Hidden State Alignment via CKA}
Low-dimensional representations possess limited capacity, making it impractical to encode exhaustive token-level details without introducing noise. We argue that forcing a small bottleneck to align with the entire sequence leads to information saturation. Instead, we use a \textbf{Top-$k$ Selection} with Centered Kernel Alignment (CKA) strategy to reorganize information, focusing exclusively on a compact subset of the most informative tokens.

\paragraph{Hard Top-$k$ Selection} For each truncated dimension $d_i$, we select the top $k_i$ tokens based on their importance scores in $a_D$, where $k_i$ increases monotonically with $d_i$. This selective strategy offers dual benefits: it significantly reduces the computational overhead of the alignment process and acts as a denoising filter, preventing low-capacity prefixes from being saturated by irrelevant or redundant context. This also creates a dual Matryoshka structure: low-dimensional cores capture a coarse semantic skeleton, while higher-dimensional spaces progressively incorporate finer details. Crucially, the selected token sets are nested: $\mathcal{S}_{k_1} \subset \mathcal{S}_{k_2} \subset \dots \subset \mathcal{S}_{k_n} \subseteq \{1, \dots, m\}$, where $\mathcal{S}_{k_i}$ denotes the indices of the $k_i$ most important tokens. This ensures that higher-capacity embeddings refine and enrich the semantic core learned by lower-capacity ones, rather than relearning it.

\paragraph{Representation Consistency via Linear CKA.}
Given these nested token subsets, we require an alignment objective that enforces structural consistency between the resulting student and teacher representations, even though they differ in feature dimensionality ($d_i$ versus $D$). To this end, we adopt Centered Kernel Alignment (CKA) with a linear kernel, which measures similarity between representation spaces in a dimension-agnostic manner. We specifically choose the linear formulation for its computational efficiency, which is critical when scaling to LLMs. Let us define two kernel matrices:

\begin{equation}
    \mathbf{p}_i =  \mathbf{h}_i\mathbf{h}_i^\top, \quad  \mathbf{P}_i = \mathbf{H}_i\mathbf{H}_i^\top
\end{equation}
We extract the submatrices for the student $\mathbf{h}_i \in \mathbb{R}^{k_i \times d_i}$ and the teacher $\mathbf{H}_i \in \mathbb{R}^{k_i \times D}$ using the nested indices defined previously. First, we compute the centered feature matrices $\tilde{\mathbf{h}}_i$ and $\tilde{\mathbf{H}}_i$ as:
\begin{equation}
    \tilde{\mathbf{h}}_i = \mathbf{h}_i(\mathbf{I}_{k_i} - \frac{1}{k_i}\mathbf{1}\mathbf{1}^\top), \quad \tilde{\mathbf{H}}_i = \mathbf{H}_i(\mathbf{I}_{k_i} - \frac{1}{k_i}\mathbf{1}\mathbf{1}^\top)
\end{equation}
where $\mathbf{I}_{k_i}$ is the identity matrix and $\mathbf{1}$ is a vector of ones. For linear kernels, the HSIC can be efficiently computed as \citep{kornblith2019similarityneuralnetworkrepresentations}:
\begin{equation}
\label{HSIC}
    \text{HSIC}(\mathbf{p}_i, \mathbf{P}_i) = \left\| \text{cov}(\tilde{\mathbf{h}}_i^\top, \tilde{\mathbf{H}}_i^\top) \right\|_F^2
\end{equation}
where $\text{cov}(\mathbf{X}, \mathbf{Y}) = \mathbf{X}^\top \mathbf{Y}$ denotes the covariance function. Moreover, from section \ref{CKA}, the Linear CKA metric between $\mathbf{h}_i$ and $\mathbf{H}_i$ is defined as:

\begin{equation}
\label{linear CKA}
\scalebox{0.9}{$
\text{CKA}\left(\mathbf{h}_i, \mathbf{H}_i\right)=
\dfrac{
    HSIC(\mathbf{p}_i, \mathbf{P}_i)
}{
    \sqrt{HSIC(\mathbf{p}_i,\mathbf{p}_i) \cdot 
    HSIC(\mathbf{P}_i, \mathbf{P}_i})
}
$}
\end{equation}
Combining two equations Eq.\ref{HSIC} and Eq.\ref{linear CKA}, we obtain the formulation of linear CKA between two hidden state matrices:

\begin{equation}
\scalebox{1}{$
\text{CKA}\left(\mathbf{h}_i, \mathbf{H}_i\right)
=  \frac{
\left\| \text{cov}(\tilde{\mathbf{h}}_i^\top, \tilde{\mathbf{H}}_i^\top) \right\|_F^2
}{
\left\| \text{cov}(\tilde{\mathbf{h}}_i^\top, \tilde{\mathbf{h}}_i^\top) \right\|_F
\cdot
\left\| \text{cov}(\tilde{\mathbf{H}}_i^\top, \tilde{\mathbf{H}}_i^\top) \right\|_F
}
$}
\end{equation}
Finally, at each $d_i$ in the Matryoshka dimensions, we define the CKA loss between the lower-dimension full-dimensional hidden state:

\begin{equation}
\mathcal{L}_{\text{CKA}}^{(i)}
= 1
- \text{CKA}\left(\mathbf{h}_i, \mathbf{H}_i\right)
\end{equation}
We minimize this loss to ensure that the lower-dimensional bottleneck learns a geometric structure to that of the full-dimensional representation space. 
By maximizing CKA on these nested subsets, we ensure that the geometric relationships and structural organization within the semantic core are preserved robustly across all granularities. We specifically adopt CKA because it provides a robust measure of representational similarity that is invariant to orthogonal transformations and isotropic scaling \citep{kornblith2019similarityneuralnetworkrepresentations}. This invariance properties of CKA is particularly critical in our framework, as we must simultaneously align multiple nested dimensions of varying capacities.

\subsubsection{Final SIA loss}

We define the composite SIA loss at layer $k$ by integrating both attention-based importance and CKA loss consistency across the entire Matryoshka dimensions $\mathcal{D} = \{d_1 < d_2 < \dots < d_n = D\}$:

\begin{equation}
    \mathcal{L}_{\text{SIA}}^{(k)} = \sum_{i=1}^{n} (\mathcal{L}_{\text{att}}^{(i)} + \mathcal{L}_{\text{CKA}}^{(i)})
\end{equation}
The total SIA loss is then computed by summing these local alignment constraints over the network's depth, specifically across the target layers $\mathcal{L}$:

\begin{equation}
    \mathcal{L}_{\text{SIA}} = \sum_{k \in \mathcal{L}} \mathcal{L}_{\text{SIA}}^{(k)}
\end{equation}

\subsection{PIC: Progressive Information Chaining}
\label{sec:pic}

Building upon the priority-based structure induced by \textbf{SIA}, which reorganizes internal representations to emphasize prefix dimensions, we introduce \textbf{PIC} to guide the flow of semantic information across network depth. While many existing MRL approaches apply task supervision primarily at the final representation, such a design places the burden of semantic compression at a single endpoint. \textbf{PIC} instead adopts a \textbf{scaffolded alignment pipeline} that provides intermediate guidance throughout the network. Motivated by the progressive specialization of Transformer layers, PIC propagates task-relevant cues from deeper layers to earlier ones, enabling lower-dimensional representations to acquire essential task semantics at early stages of learning. Supervision is restricted to the compact subspace reorganized by SIA, concentrating alignment on the most informative dimensions while preserving low-level features and representational flexibility in the remaining space.

\paragraph{Formal Objective.}
Let the scaffolding structure be defined by a sequence of $n$ checkpoints $\mathcal{C} =\{(l_i, d_i)\}_{i=1}^n$, ordered by depth and capacity such that $l_1 < l_2 < \dots < l_n$ and $d_1 < d_2 < \dots < d_n$. Here, $l_i$ denotes the layer index and $d_i$ the embedding dimension. We denote the truncated \texttt{[CLS]} representation at checkpoint $i$ as $\mathbf{z}_{i} \in \mathbb{R}^{d_i}$.

Our aim is to promote a progressive condensation of task-relevant information, such that each checkpoint learns to supply the essential signals required by the next, allowing early, low-capacity representations to serve as reliable building blocks for deeper ones. Intuitively, this can be framed as encouraging high mutual information (MI) between adjacent checkpoints: maximizing \(I(\mathbf{z}_i; \mathbf{z}_{i+1})\) requires that the earlier representation \(\mathbf{z}_i\) retain the core features needed to predict the next representation \(\mathbf{z}_{i+1}\). In the context of Matryoshka learning this enforces a progressive condensation of task-relevant signals rather than a single long semantic jump from early checkpoints to the final output. Since calculating exact MI is computationally impossible, we use \textbf{InfoNCE} as an approximation:

\begin{equation}
    I(\mathbf{z}_i; \mathbf{z}_{i+1}) \geq \log N - \mathcal{L}_{\text{InfoNCE}}
\end{equation}
where $N$ is the batch size. To bridge the dimensionality mismatch ($d_i \neq d_{i+1}$), we employ a lightweight nonlinear projector $\phi_i: \mathbb{R}^{d_i} \to \mathbb{R}^{d_{i+1}}$ to map the lower-dimensional upstream representation into the semantic space of the downstream target. The local alignment loss for each step is:
\begin{equation}
\resizebox{1\linewidth}{!}{$
\mathcal{L}_{\text{chain}}^{(i)} = - \mathbb{E} \left[
\log \frac{\exp(\text{sim}(\phi_i(\mathbf{z}_i), \mathbf{z}_{i+1}) / \tau)}
{\sum_{\mathbf{z}' \in \mathcal{B}_{i+1}}
\exp(\text{sim}(\phi_i(\mathbf{z}_i), \mathbf{z}') / \tau)}
\right]
$}
\end{equation}
where $\mathcal{B}_{i+1}$ contains the positive pair $\mathbf{z}_{i+1}$ and $N-1$ negative samples from the batch, and $\tau$ is the temperature scaling parameter. The total Progressive Information Chaining loss accumulates these local alignment signals:
\begin{equation}
    \mathcal{L}_{\text{PIC}} = \sum_{i=1}^{n-1} \mathcal{L}_{\text{chain}}^{(i)}
\end{equation}
This chain-based alignment fundamentally differs from direct supervision by focusing on \textbf{essential information transfer}. By aligning only the most salient parts of adjacent layers via Mutual Information, PIC excessive information loss while avoiding the rigidity of element-wise matching. It ensures that the early layers act as a flexible foundation, organizing task-relevant features in a way that supports, rather than strictly replicates, the subsequent refinement stages. This maintains the natural diversity of features across depth, leading to a more robust final representation.

\subsection{Overall Training Objective}
The full objective of MIPIC integrates our proposed self-distillation mechanisms with the standard Matryoshka formulation. The total loss is:
\begin{equation}
\resizebox{0.48\textwidth}{!}{
    $
    \mathcal{L}_{\text{MIPIC}}
    = \alpha\mathcal{L}_{\text{MRL}}
    + (1-\alpha)\!\left[
    \mathcal{L}_{\text{SIA}}
    + \mathcal{L}_{\text{PIC}}
    \right]
    $
}
\end{equation}
Here, $\mathcal{L}_{\text{MRL}}$ denotes the Matryoshka task-specific loss (like in Equation \ref{MRL loss}). One hyperparameter is used: $\alpha \in [0,1]$ trades off the MRL and MIPIC.

\begin{table*}[h!]
\centering
\resizebox{\textwidth}{!}{%
% Cấu trúc cột: l c | cccc | cccc (Ngắt đôi đậm) l c | cccc | cccc
\begin{tabular}{lc|cccc|cccc!{\vrule width 1.5pt}lc|cccc|cccc}
\toprule
\multirow{2}{*}{\textbf{Datasets}} & \multirow{2}{*}{\textbf{Dim.}} & \multicolumn{4}{c|}{\textbf{TinyBERT 6L}} & \multicolumn{4}{c!{\vrule width 1.5pt}}{\textbf{BERT}} & 
\multirow{2}{*}{\textbf{Datasets}} & \multirow{2}{*}{\textbf{Dim.}} & \multicolumn{4}{c|}{\textbf{TinyBERT 6L}} & \multicolumn{4}{c}{\textbf{BERT}} \\
\cmidrule(lr){3-6} \cmidrule(lr){7-10} \cmidrule(lr){13-16} \cmidrule(lr){17-20}
 & & \small{Unsup SimCSE} & \small{MRL} & \small{ESE} & \small{MIPIC} & \small{Unsup SimCSE} & \small{MRL} & \small{ESE} & \small{MIPIC} & 
 & & \small{Unsup SimCSE} & \small{MRL} & \small{ESE} & \small{MIPIC} & \small{Unsup SimCSE} & \small{MRL} & \small{ESE} & \small{MIPIC} \\
\midrule

% Block 1: Banking77 & WIC
\multirow{7}{*}{Banking77} 
 & 16 & 30.17 & 40.64 & 43.30 & \textbf{48.93} & 35.92 & 46.39 & 45.86 & \textbf{54.65} & \multirow{7}{*}{WIC} 
 & 16 & 43.90 & 58.07 & 58.12 & \textbf{60.78} & 48.46 & 60.13 & 61.14 & \textbf{62.42} \\
 & 32 & 56.23 & 63.48 & 63.78 & \textbf{65.68} & 54.23 & 64.90 & 66.39 & \textbf{71.67} & 
 & 32 & 45.86 & 59.85 & 59.35 & \textbf{61.71} & 55.11 & 62.42 & 62.64 & \textbf{63.28} \\
 & 64 & 66.10 & \textbf{74.89} & 73.96 & 74.88 & 67.78 & 76.84 & 79.04 & \textbf{79.23} & 
 & 64 & 51.14 & 61.35 & 59.95 & \textbf{62.72} & 57.20 & 64.28 & 63.57 & \textbf{64.53} \\
 & 128 & 75.32 & 81.97 & 81.53 & \textbf{82.43} & 75.43 & 83.49 & 86.06 & \textbf{86.24} & 
 & 128 & 59.27 & 62.78 & 60.42 & \textbf{63.50} & 63.17 & 65.24 & 66.14 & \textbf{66.64} \\
 & 256 & 85.82 & 85.51 & 85.76 & \textbf{86.78} & 85.84 & 86.45 & 87.59 & \textbf{87.64} & 
 & 256 & 61.29 & 63.71 & 60.76 & \textbf{64.50} & 62.93 & 66.42 & 66.92 & \textbf{66.98} \\
 & 512 & 87.78 & 87.20 & 87.04 & \textbf{88.15} & 88.34 & 87.85 & 88.93 & \textbf{89.22} & 
 & 512 & 62.23 & 63.14 & 61.50 & \textbf{64.01} & 65.78 & 65.85 & \textbf{67.77} & 67.12 \\
 & 768 & 88.29 & 87.70 & \textbf{88.89} & 88.76 & 89.15 & 88.43 & \textbf{89.83} & 89.45 & 
 & 768 & 63.21 & 63.07 & 61.85 & \textbf{64.37} & 65.71 & 65.91 & \textbf{67.42} & 67.28 \\
\midrule % Đường kẻ ngang phân cách dataset

% Block 2: TweetEval & SICK
\multirow{7}{*}{TweetEval} 
 & 16 & 35.12 & 53.66 & 52.71 & \textbf{60.57} & 48.85 & 55.96 & 50.73 & \textbf{60.38} & \multirow{7}{*}{SICK} 
 & 16 & 39.27 & 59.07 & 58.42 & \textbf{61.03} & 45.75 & 59.35 & 61.50 & \textbf{62.07} \\
 & 32 & 46.06 & 59.08 & 59.17 & \textbf{63.32} & 55.50 & 60.51 & 54.75 & \textbf{62.30} & 
 & 32 & 51.06 & 65.08 & 64.98 & \textbf{67.38} & 51.86 & 63.34 & 62.36 & \textbf{64.24} \\
 & 64 & 49.72 & 60.22 & 65.16 & \textbf{65.36} & 63.27 & 65.68 & 60.55 & \textbf{65.64} & 
 & 64 & 55.38 & \textbf{68.47} & 66.92 & 67.32 & 52.97 & 66.11 & 62.07 & \textbf{66.26} \\
 & 128 & 55.89 & 66.71 & 66.75 & \textbf{67.61} & 66.23 & 68.14 & 65.51 & \textbf{68.24} & 
 & 128 & 61.78 & \textbf{69.84} & 68.73 & 68.46 & 59.33 & 66.57 & 66.74 & \textbf{67.00} \\
 & 256 & 63.54 & 68.03 & \textbf{69.93} & 69.29 & 68.12 & 69.38 & 68.73 & \textbf{70.45} & 
 & 256 & 68.93 & 68.85 & 69.03 & \textbf{70.82} & 65.72 & 67.23 & 66.43 & \textbf{67.44} \\
 & 512 & 66.47 & 68.25 & 69.32 & \textbf{70.43} & 70.02 & 70.86 & 70.22 & \textbf{71.58} & 
 & 512 & 70.46 & 70.09 & 70.57 & \textbf{70.80} & 67.41 & \textbf{68.33} & 67.04 & 68.02 \\
 & 768 & 70.16 & 69.10 & 70.02 & \textbf{70.89} & 70.22 & 70.72 & 70.75 & \textbf{71.26} & 
 & 768 & 70.55 & 70.00 & \textbf{70.67} & 70.65 & 68.15 & \textbf{68.94} & 68.19 & 68.73 \\
\midrule

% Block 3: MRPC & STSB
\multirow{7}{*}{MRPC} 
 & 16 & 51.32 & 61.12 & 65.12 & \textbf{72.34} & 57.83 & 65.93 & 66.73 & \textbf{72.46} & \multirow{7}{*}{STSB} 
 & 16 & 45.63 & 60.00 & 57.45 & \textbf{62.43} & 49.24 & 61.65 & 59.39 & \textbf{63.43} \\
 & 32 & 57.34 & 69.87 & 69.38 & \textbf{74.02} & 66.72 & 70.11 & 71.07 & \textbf{73.33} & 
 & 32 & 58.78 & 63.51 & 60.12 & \textbf{65.47} & 51.34 & 65.09 & 64.39 & \textbf{66.16} \\
 & 64 & 68.25 & 72.28 & 70.12 & \textbf{74.55} & 70.53 & 70.14 & 72.57 & \textbf{73.68} & 
 & 64 & 58.37 & 65.45 & 65.78 & \textbf{67.37} & 61.22 & 67.32 & 67.51 & \textbf{68.02} \\
 & 128 & 68.15 & 72.34 & 72.89 & \textbf{73.97} & 73.56 & 72.96 & 71.71 & \textbf{74.02} & 
 & 128 & 60.83 & 67.16 & 68.52 & \textbf{68.81} & 67.53 & 69.59 & 70.12 & \textbf{70.55} \\
 & 256 & 71.82 & 72.52 & 73.97 & \textbf{74.37} & 73.73 & 72.98 & 72.29 & \textbf{74.14} & 
 & 256 & 66.61 & 67.79 & 68.36 & \textbf{69.18} & 68.23 & 70.41 & \textbf{70.91} & 70.10 \\
 & 512 & 72.16 & 72.76 & 74.31 & \textbf{74.67} & 74.08 & 73.16 & 73.27 & \textbf{74.62} & 
 & 512 & 67.82 & 68.85 & 68.92 & \textbf{69.54} & 70.67 & \textbf{71.33} & 70.83 & 71.22 \\
 & 768 & 72.51 & 72.87 & 74.02 & \textbf{74.60} & 74.02 & 73.22 & 72.46 & \textbf{74.44} & 
 & 768 & 69.45 & 68.93 & 69.02 & \textbf{69.74} & 71.58 & 71.30 & \textbf{71.99} & 71.96 \\
\bottomrule
\end{tabular}%
}
\caption{Results on in-domain datasets with TinyBERT 6L and BERT backbone embedding models. Bold indicates the best result at the same representation size, backbone model, and dataset.}
\label{tab1}
\end{table*}

\begin{table*}[h!]
\centering
\resizebox{\textwidth}{!}{%
\begin{tabular}{lc|cccc|cccc!{\vrule width 1.5pt}lc|cccc|cccc}
\toprule
\multirow{2}{*}{\textbf{Datasets}} & \multirow{2}{*}{\textbf{Dim.}} & \multicolumn{4}{c|}{\textbf{TinyBERT 6L}} & \multicolumn{4}{c!{\vrule width 1.5pt}}{\textbf{BERT}} & 
\multirow{2}{*}{\textbf{Datasets}} & \multirow{2}{*}{\textbf{Dim.}} & \multicolumn{4}{c|}{\textbf{TinyBERT 6L}} & \multicolumn{4}{c}{\textbf{BERT}} \\
\cmidrule(lr){3-6} \cmidrule(lr){7-10} \cmidrule(lr){13-16} \cmidrule(lr){17-20}
 & & \small{Unsup SimCSE} & \small{MRL} & \small{ESE} & \small{MIPIC} & \small{Unsup SimCSE} & \small{MRL} & \small{ESE} & \small{MIPIC} & 
 & & \small{Unsup SimCSE} & \small{MRL} & \small{ESE} & \small{MIPIC} & \small{Unsup SimCSE} & \small{MRL} & \small{ESE} & \small{MIPIC} \\
\midrule

% Block 1: STS12 & STS16
\multirow{7}{*}{STS12} 
 & 16 & 45.68 & 51.48 & 50.90 & \textbf{59.34} & 47.88 & 55.13 & 51.34 & \textbf{62.08} & \multirow{7}{*}{STS16} 
 & 16 & 51.26 & 59.21 & 60.26 & \textbf{61.05} & 50.78 & 54.78 & 59.67 & \textbf{64.32} \\
 & 32 & 47.09 & 55.25 & 56.02 & \textbf{61.60} & 53.84 & 59.78 & 53.78 & \textbf{64.87} & 
 & 32 & 53.46 & 63.45 & 65.63 & \textbf{66.23} & 53.34 & 60.96 & 64.54 & \textbf{68.00} \\
 & 64 & 53.48 & 55.67 & 58.32 & \textbf{62.70} & 61.22 & 61.24 & 55.23 & \textbf{65.88} & 
 & 64 & 60.72 & 66.29 & 66.04 & \textbf{68.43} & 57.23 & 63.45 & 68.15 & \textbf{69.32} \\
 & 128 & 59.50 & 59.80 & 60.38 & \textbf{63.74} & 65.45 & 62.34 & 59.97 & \textbf{66.32} & 
 & 128 & 66.98 & 67.90 & 68.12 & \textbf{69.86} & 65.89 & 66.78 & 70.14 & \textbf{70.30} \\
 & 256 & 61.14 & 60.17 & 61.23 & \textbf{63.82} & 65.83 & 64.13 & 60.33 & \textbf{67.08} & 
 & 256 & 67.25 & 68.10 & \textbf{70.55} & 69.92 & 66.86 & 69.89 & 70.82 & \textbf{70.86} \\
 & 512 & 61.77 & 61.52 & 60.82 & \textbf{64.14} & 66.16 & 65.09 & 61.06 & \textbf{67.55} & 
 & 512 & 68.94 & 68.60 & \textbf{70.87} & 70.57 & 70.25 & 70.60 & \textbf{71.47} & 70.78 \\
 & 768 & 62.19 & 61.78 & 64.16 & \textbf{64.57} & 66.31 & 64.84 & 60.43 & \textbf{67.64} & 
 & 768 & 69.26 & 68.69 & 70.25 & \textbf{70.93} & 71.07 & 70.53 & 70.54 & \textbf{71.56} \\
\midrule

% Block 2: STS13 & SickR
\multirow{7}{*}{STS13} 
 & 16 & 49.58 & 55.03 & 53.51 & \textbf{64.30} & 55.34 & 62.13 & 60.48 & \textbf{65.39} & \multirow{7}{*}{SickR} 
 & 16 & 57.35 & 60.43 & 60.79 & \textbf{61.17} & 55.67 & 63.45 & 64.06 & \textbf{64.78} \\
 & 32 & 50.60 & 63.48 & 60.46 & \textbf{67.51} & 61.23 & 63.44 & 65.78 & \textbf{68.81} & 
 & 32 & 66.01 & 66.71 & 66.88 & \textbf{67.45} & 57.23 & 65.92 & 66.77 & \textbf{67.34} \\
 & 64 & 58.43 & 62.96 & 64.56 & \textbf{68.57} & 67.88 & 68.98 & 68.97 & \textbf{71.22} & 
 & 64 & 68.43 & \textbf{68.45} & 68.30 & 67.93 & 60.86 & 68.25 & 69.09 & \textbf{70.89} \\
 & 128 & 60.93 & 67.88 & 66.40 & \textbf{68.73} & 71.09 & 70.97 & 71.68 & \textbf{71.87} & 
 & 128 & 70.02 & 69.82 & 69.39 & \textbf{70.73} & 64.23 & 69.61 & 70.30 & \textbf{70.43} \\
 & 256 & 61.24 & 67.80 & 68.53 & \textbf{69.84} & 71.88 & 71.24 & 72.32 & \textbf{72.62} & 
 & 256 & 70.20 & 69.87 & 69.63 & \textbf{70.55} & 66.78 & 70.49 & 71.08 & \textbf{72.56} \\
 & 512 & 67.92 & 68.96 & 69.65 & \textbf{70.22} & 72.13 & 72.35 & \textbf{73.63} & 73.33 & 
 & 512 & 70.49 & 70.16 & 70.38 & \textbf{71.03} & 67.34 & 70.83 & 71.48 & \textbf{71.78} \\
 & 768 & 70.54 & 69.16 & 69.94 & \textbf{70.87} & 74.78 & 73.56 & 73.81 & 73.49 & 
 & 768 & 70.58 & 70.10 & 70.49 & \textbf{71.94} & 67.89 & 70.98 & 71.58 & \textbf{72.56} \\
\midrule

% Block 3: STS14 & Emotion
\multirow{7}{*}{STS14} 
 & 16 & 49.35 & 50.42 & 50.12 & \textbf{57.27} & 50.67 & 51.76 & 54.20 & \textbf{56.93} & \multirow{7}{*}{Emotion} 
 & 16 & 28.44 & 31.30 & 29.76 & \textbf{32.87} & 24.78 & 26.98 & 27.95 & \textbf{30.24} \\
 & 32 & 48.23 & 53.67 & 53.27 & \textbf{60.09} & 57.23 & 54.78 & 56.89 & \textbf{61.45} & 
 & 32 & 32.65 & 37.94 & 36.11 & \textbf{41.56} & 34.15 & 36.87 & 37.49 & \textbf{41.55} \\
 & 64 & 51.28 & 59.57 & 57.64 & \textbf{61.32} & 60.86 & 60.88 & 60.44 & \textbf{63.13} & 
 & 64 & 41.35 & 43.53 & 44.29 & \textbf{51.15} & 42.46 & 42.52 & 44.61 & \textbf{50.97} \\
 & 128 & 55.98 & 60.84 & 61.42 & \textbf{62.15} & 61.25 & 62.05 & 61.45 & \textbf{63.97} & 
 & 128 & 46.39 & 49.50 & 52.37 & \textbf{55.25} & 49.76 & 49.15 & 50.37 & \textbf{53.43} \\
 & 256 & 60.13 & 60.91 & 62.36 & \textbf{62.84} & 61.56 & 63.68 & 63.81 & \textbf{64.89} & 
 & 256 & 52.53 & 53.92 & 56.12 & \textbf{57.55} & 54.42 & \textbf{54.67} & 54.33 & 53.75 \\
 & 512 & \textbf{63.93} & 62.22 & 63.47 & 63.22 & 62.45 & 64.56 & 64.63 & \textbf{65.43} & 
 & 512 & 54.93 & 56.39 & 56.82 & \textbf{60.87} & 54.36 & 55.98 & 56.63 & \textbf{57.98} \\
 & 768 & 63.01 & 62.32 & \textbf{63.85} & 63.31 & 65.58 & 64.65 & 64.85 & 65.37 & 
 & 768 & 55.30 & 58.53 & \textbf{60.43} & 60.39 & 53.78 & 54.08 & 57.42 & \textbf{58.65} \\
\midrule

% Block 4: STS15 & SciTail
\multirow{7}{*}{STS15} 
 & 16 & 60.92 & 64.17 & 63.84 & \textbf{66.85} & 61.45 & 61.99 & 64.78 & \textbf{68.85} & \multirow{7}{*}{SciTail} 
 & 16 & 65.34 & 72.25 & 71.87 & \textbf{73.97} & 66.34 & 68.45 & 69.61 & \textbf{72.48} \\
 & 32 & 65.45 & 69.88 & 69.23 & \textbf{71.05} & 65.34 & 70.62 & 69.46 & \textbf{71.35} & 
 & 32 & 66.45 & 74.92 & 73.32 & \textbf{75.54} & 70.12 & 68.78 & 73.51 & \textbf{74.01} \\
 & 64 & 69.32 & 72.51 & 72.08 & \textbf{73.20} & 65.78 & 73.26 & 72.64 & \textbf{74.47} & 
 & 64 & 70.73 & 75.40 & 73.75 & \textbf{75.67} & 71.23 & 71.34 & \textbf{75.02} & 73.96 \\
 & 128 & 70.54 & 73.24 & 73.75 & \textbf{74.41} & 66.23 & 75.28 & \textbf{75.63} & 74.72 & 
 & 128 & 71.88 & 75.68 & 74.90 & \textbf{75.78} & 74.78 & 72.56 & 75.77 & \textbf{75.89} \\
 & 256 & 74.76 & 73.29 & \textbf{76.12} & 75.21 & 71.32 & 76.24 & \textbf{76.93} & 75.94 & 
 & 256 & 72.45 & 75.92 & 75.53 & \textbf{75.97} & 76.95 & 75.77 & 76.29 & \textbf{77.24} \\
 & 512 & \textbf{75.38} & 73.87 & 75.24 & 76.23 & 74.56 & \textbf{76.73} & 76.12 & 76.38 & 
 & 512 & 72.84 & \textbf{75.82} & 75.77 & 75.34 & 76.66 & 75.72 & 76.38 & \textbf{76.89} \\
 & 768 & 75.43 & 73.90 & 76.24 & \textbf{76.34} & \textbf{77.39} & 76.62 & 76.94 & 76.22 & 
 & 768 & 75.54 & \textbf{75.90} & 75.72 & 75.61 & 76.57 & 76.34 & \textbf{76.85} & 76.54 \\
\bottomrule
\end{tabular}%
}
\caption{Results on out-domain datasets with TinyBERT 6L and BERT backbone embedding models. Bold indicates the best result at the same representation size, backbone model, and dataset.}
\label{tab2}
\end{table*}

\section{Experiments}
\subsection{Experimental Setup}

We evaluate embeddings across in-domain and out-of-domain settings to assess both task performance and generalization.
\paragraph{Task categories.} We report results across three representative task families, all of which rely critically on embedding quality. First, for \textbf{Text Classification}, which requires models to capture the overall semantics of a single text input, we evaluate on TweetEval \citep{barbieri-etal-2020-tweeteval}, as well as Emotion and Banking77 from the MTEB benchmark \citep{muennighoff2023mtebmassivetextembedding}. Second, regarding \textbf{NLI}, which focuses on modeling semantic relationships between two inputs, we consider MRPC from GLUE \citep{wang-etal-2018-glue}, WiC \citep{pilehvar-camacho-collados-2019-wic}, and SciTail \citep{scitail}. Finally, for \textbf{Semantic Textual Similarity (STS)}, which measures the ability to capture fine-grained semantic similarity, we evaluate on STS-B and SICK \citep{marelli-etal-2014-semeval} for in-domain testing, and extend to STS12--16 and SickR \citep{muennighoff2023mtebmassivetextembedding} for OOD evaluation. Additional details regarding model architectures, training procedures, datasets, and evaluation protocols are provided in Appendix~\ref{sec:appendix_experiments}.

\paragraph{Baselines.}
We compare against SOTA Matryoshka baselines: \textbf{MRL} \citep{NEURIPS2022_c32319f4}, which learns nested prefixes via multi-scale supervision, and \textbf{ESE} \citep{li2025ese}, which scales across both dimensionality and model depth. These provide a strong benchmark for evaluating low-dimensional representation quality.

\subsection{Results}

Results for in-domain and out-of-domain benchmarks (Table \ref{tab1} and Table \ref{tab2}) indicate that MIPIC achieves competitive or superior performance compared to baselines, including Unsup SimCSE, MRL, and Espresso. While maintaining comparable efficacy at higher dimensions, the performance gap becomes particularly distinct at extremely low dimensions (16 and 32), demonstrating the framework's superior capability in semantic compression. These trends hold consistent across TinyBERT and BERT architectures. Furthermore, scalability evaluations on larger backbones such as BGEM3 and Qwen3 0.6B (Appendix \ref{more ablation}) validate the method's robustness, confirming that while MIPIC maintains parity with baselines in full-capacity regimes, it significantly excels in retaining semantic density under extreme truncation.

\section{Analysis}

We analyze the individual contributions of SIA and PIC, alongside the effectiveness of our progressive dimension scaling strategy to validate MIPIC's design.

\paragraph{Impact of Framework Components} Table~\ref{tab:ablation} highlights the synergy between SIA and PIC. SIA is vital for structural organization; its removal causes sharp drops at small sizes, proving geometric order is key for information density. Meanwhile, PIC acts as a semantic bridge to stabilize task knowledge across layers. While SIA ensures geometric efficiency, PIC prevents semantic loss in early stages. Together, they enable a coarse-to-fine refinement that maintains high quality at every dimension.
\begin{table}[htbp]
\centering
\resizebox{\columnwidth}{!}{% 
\begin{tabular}{l c ccc}
\toprule
\textbf{Datasets} & \textbf{Rep. size} & \textbf{MIPIC} & \textbf{MIPIC w/o SIA} & \textbf{MIPIC w/o PIC} \\
\midrule
\multirow{7}{*}{STS12} 
 & 16  & \textbf{62.08} & 60.12 & 61.27 \\
 & 32  & \textbf{64.87} & 62.39 & 63.90 \\
 & 64  & \textbf{65.88} & 64.82 & 65.54 \\
 & 128 & \textbf{66.32} & 65.34 & 65.72 \\
 & 256 & \textbf{67.08} & 66.87 & 66.97 \\
 & 512 & \textbf{67.55} & 66.43 & 67.12 \\
 & 768 & \textbf{67.64} & 67.12 & 67.53 \\
\midrule
\multirow{7}{*}{TweetEval} 
 & 16  & \textbf{60.38} & 57.12 & 59.13 \\
 & 32  & \textbf{62.30} & 59.32 & 61.87 \\
 & 64  & \textbf{65.64} & 65.02 & 65.11 \\
 & 128 & \textbf{68.24} & 67.35 & 67.92 \\
 & 256 & \textbf{70.45} & 69.54 & 70.12 \\
 & 512 & \textbf{71.58} & 70.24 & 71.55 \\
 & 768 & \textbf{71.26} & 70.22 & 70.37 \\
\bottomrule
\end{tabular}
}
\caption{Ablation study results showing the impact of SIA and PIC. ``w/o'' denotes ``without''.}
\label{tab:ablation}
\end{table}

\paragraph{Effectiveness of Progressive Dimension Design} Table~\ref{tab:pic_ablation} shows that progressively increasing dimensions works better than keeping all layers at full size. Using full dimensions everywhere removes the bottleneck needed for Matryoshka learning. Our design forces early layers to pack the most important information into small prefixes, then refine it later. This leads to more stable and effective representations. Without this constraint, the model does not learn to focus on key information early, which hurts performance.

\begin{table}[ht]
\centering
\resizebox{\columnwidth}{!}{%
\begin{tabular}{l c cc}
\toprule
\textbf{Datasets} & \textbf{Rep. size} & \textbf{Our design} & \textbf{All dim equal} \\
\midrule
\multirow{7}{*}{STS12} 
 & 16  & \textbf{62.08} & 61.98 \\
 & 32  & \textbf{64.87} & 63.12 \\
 & 64  & \textbf{65.88} & 64.77 \\
 & 128 & \textbf{66.32} & 66.12 \\
 & 256 & \textbf{67.08} & 66.52 \\
 & 512 & \textbf{67.55} & 67.12 \\
 & 768 & \textbf{67.64} & 67.56 \\
\midrule
\multirow{7}{*}{Tweet Eval} 
 & 16  & \textbf{60.38} & 58.99 \\
 & 32  & \textbf{62.30} & 61.97 \\
 & 64  & \textbf{65.64} & 65.33 \\
 & 128 & \textbf{68.24} & 67.92 \\
 & 256 & \textbf{70.45} & 70.08 \\
 & 512 & \textbf{71.58} & 71.26 \\
 & 768 & \textbf{71.26} & 71.15 \\
\bottomrule
\end{tabular}
}
\caption{Ablation study on PIC design. ``Our design'' uses progressive dimension scaling across layers, while ``All dim equal'' maintains full dimensions at every stage.}
\label{tab:pic_ablation}
\end{table}

\paragraph{Training time analysis}
We explicitly analyze the computational cost of our framework compared to the MRL and ESE baselines using the BERT-base backbone. The training throughput (measured in iterations per second and samples per second) is reported in Table \ref{tab:training_time}.
\begin{table}[htbp]
    \centering
    \resizebox{\linewidth}{!}{
    \begin{tabular}{l|cc}
        \toprule
        \textbf{Method} & \textbf{Iterations / s} & \textbf{Throughput (sample/s)} \\
        \midrule
        MRL & 7.61 & 121.76 \\
        ESE & 5.24 & 83.84 \\
        \textbf{MIPIC (Ours)} & 2.88 & 46.08 \\
        \bottomrule
    \end{tabular}
    }
    \caption{Training efficiency comparison on the BERT-base backbone. MIPIC incurs higher training latency due to the computation of auxiliary alignment losses (SIA and PIC).}
    \label{tab:training_time}
\end{table}
As indicated in Table \ref{tab:training_time}, MIPIC exhibits a lower training throughput compared to baselines. Specifically, our method operates at approximately $2.88$ iterations/s, representing a reduction of roughly $45\%$ compared to ESE ($5.24$ it/s) and $62\%$ compared to the standard MRL ($7.61$ it/s). This increased overhead is expected, as MIPIC requires additional forward passes and gradient computations for the multi-layer alignment objectives (SIA and PIC) involving CKA and InfoNCE calculations. However, it is crucial to distinguish between \textit{training cost} and \textit{inference latency}. The computational overhead shown above is strictly a \textbf{one-time training investment}. Structurally, MIPIC does not introduce any additional parameters or modules to the backbone encoder during inference. All auxiliary projectors (e.g., $P_i$ in SIA, $\phi_i$ in PIC) are discarded after training. Consequently, a deployed MIPIC model possesses \textbf{zero additional inference latency} compared to a standard backbone model. Given that representation models are typically trained once but queried millions of times, we argue that the increased training time is a justifiable trade-off for the significant gains in representation quality and compression efficiency demonstrated in our main experiments.

\section{Conclusion}

We proposed MIPIC, a framework that coordinates Matryoshka Representation Learning across both embedding dimensions and network depth. By combining Self-Distilled Intra-Relational Alignment (SIA) for structural consistency and Progressive Information Chaining (PIC) for semantic consolidation, MIPIC establishes a global chaining of task signals. Extensive experiments on models ranging from TinyBERT to Qwen3 and BGE-M3 show that MIPIC remains robustly competitive with state-of-the-art baselines at full capacity, while achieving superior efficacy in extreme 16 and 32-dimensional compression.

\section{Limitations}

Despite its effectiveness, MIPIC introduces certain limitations, primarily the increased computational overhead during the training phase due to the calculation of auxiliary CKA and InfoNCE losses across multiple layers. The framework’s performance also relies on the careful tuning of hyperparameters and the strategic selection of layer checkpoints, which may require additional optimization when adapting to backbone architectures with different depths. Furthermore, while we validated MIPIC on a wide array of discriminative NLP tasks using encoder-based models, its utility in decoder-only generative scenarios or multi-modal settings has not yet been explored. Lastly, our current progressive dimension scaling design represents a specific configuration for depth-wise transfer, and alternative scheduling methods for information condensation could potentially yield different results.

\section*{Acknowledgments}
Trung Le was supported by the Air Force Office of Scientific Research under award number FA2386-25-1-4023 and the ARC Discovery Project grant DP250100262.

\bibliography{ref}
\bibliographystyle{acl_natbib}

\appendix

\section{Experimental Details}
\label{sec:appendix_experiments}

\subsection{Model Architectures}
To evaluate the scalability and generalizability of the MIPIC framework, we conduct experiments across a diverse range of backbone architectures varying in size and complexity. These include the compact \textbf{TinyBERT-6L} with 6 Transformer layers, the standard 12-layer \textbf{BERT-base} encoder, and modern large-scale models such as the \textbf{Qwen3-0.6B embedding} and the high-performance \textbf{BGE-M3}. By spanning from small-scale encoders to large-scale language model embeddings, we demonstrate that MIPIC consistently optimizes the internal organization of semantic information regardless of the model's depth or total parameter count.

\subsection{Detailed Dataset Statistics}

\begin{table*}[htbp]
\centering
\caption{Dataset Statistics for training and test set}
\label{tab:dataset-statistics}
\resizebox{0.5\textwidth}{!}{%
\begin{tabular}{lcc}
\toprule
\textbf{Dataset} & \textbf{Train (Sampled)} & \textbf{Test Size} \\
\midrule
Banking77 & 3,000 & 3,080 \\
TweetEval & 3,000 & 3434  \\
Emotion (OOD) & - & 1,990  \\
MRPC & 1,500 & 1,730  \\
WiC & 1,500 & 1,400 \\
SciTail (OOD) & - & 2,130 \\
SICK & 1,500 & 4,823 \\
STS-B & 1,500 & 1,390 \\
STS12 (OOD) & - & 3,108 \\
STS13 (OOD) & - & 1,500 \\
STS14 (OOD) & - & 3,750 \\
STS15 (OOD) & - & 3,000 \\
STS16 (OOD) & - & 1,186 \\
SickR (OOD) & - & 9,927 \\
\bottomrule
\end{tabular}}
\end{table*}
We use a diverse collection of datasets for both training and evaluation, covering text classification, natural language inference (NLI), and semantic textual similarity (STS). To build training data for contrastive sentence representation learning, we sample sentences from multiple task categories to promote domain and objective diversity. Specifically, we collect 6,000 sentences from classification datasets (3,000 per dataset), 3,000 sentence pairs from STS datasets (1,500 per dataset), and 3,000 sentence pairs from pair classification datasets (1,500 per dataset). All sentence pairs are flattened into individual sentences, resulting in 24,000 unique sentences, which are then used to train the backbone encoder with unsupervised SimCSE-style contrastive learning under a unified training framework. For evaluation, we test the trained models on both in-domain test sets and unseen out-of-domain datasets, including Emotion, SciTail, and multiple STS benchmarks (STS12–STS16 and SickR), which differ from the training data in domain, style, and annotation protocol. Dataset statistics are reported in Table~\ref{tab:dataset-statistics}, and all baseline methods (MRL, ESE) are trained on the same training corpus and evaluated on the same set of test datasets for fair comparison

\subsection{Training Configurations}
The detailed training configurations for the MIPIC framework across our various backbones are summarized in Table~\ref{tab:training-config}. We maintain a consistent setup to ensure a fair comparison across different scales, applying specific hyperparameters like $\alpha$ to trade off MRL and MIPIC losses. We explored the loss balancing hyperparameter $\alpha$ over the set $\{0.1, 0.2, 0.3,0.4, 0.5,0.6, 0.7\}$. The optimal configurations for each backbone are also reported in Table~\ref{tab:training-config}.

\begin{table*}[ht]
\centering
\caption{Detailed training configurations for MIPIC across different backbones.}
\label{tab:training-config}
\begin{tabular}{lcccc}
\toprule
\textbf{Configuration} & \textbf{TinyBERT-6L} & \textbf{BERT-base} & \textbf{Qwen3-0.6B} & \textbf{BGE-M3} \\
\midrule
Epochs & 5 & 5 & 5 & 5 \\
Learning Rate & $2 \times 10^{-5}$ & $2 \times 10^{-5}$ & $2 \times 10^{-5}$ & $2 \times 10^{-5}$ \\
Max Length & 256 & 256 & 256 & 256 \\
Batch Size & 16 & 16 & 16 & 16 \\
LR Scheduler & Cosine & Cosine & Cosine & Cosine \\
Optimizer & AdamW & AdamW & AdamW & AdamW \\
Temperature ($\tau$) & 0.05 & 0.05 & 0.05 & 0.05 \\
$\alpha$ (MRL weight) & 0.4 & 0.4 & 0.5 & 0.5 \\
\bottomrule
\end{tabular}
\end{table*}

\paragraph{Task Objective Specification.} In our experimental setting, the task-specific loss component within the MRL objective (Eq. \ref{MRL loss}) is instantiated as $\mathcal{L}_{\text{SimCSE}}$, the unsupervised contrastive loss adopted from \citep{gao2022simcsesimplecontrastivelearning} . Given a batch of input sentences, we feed them through the student encoder twice with different standard dropout masks $z, z'$ to obtain two views of embeddings $e_i^z$ and $e_i^{z'}$. The loss is formulated as:
\begin{equation}
    \mathcal{L}_{\text{SimCSE}} = -\frac{1}{N} \sum_{i=1}^{N} \log \frac{e^{\text{sim}(e_i^z, e_i^{z'}) / \tau}}{\sum_{j=1}^{N} e^{\text{sim}(e_i^z, e_j^{z'}) / \tau}},
    \label{eq:simcse}
\end{equation}
where $N$ is the batch size, $\tau$ is the temperature hyperparameter, and $\text{sim}(\cdot)$ denotes cosine similarity.

\subsection{Evaluation}

To assess the efficacy of the learned sentence embeddings, we conduct evaluations across three distinct categories of downstream tasks. First, for classification tasks, we follow the standard protocol established by \citep{conneau2018sentevalevaluationtoolkituniversal}, which involves training a Logistic Regression classifier on top of frozen sentence representations. Second, in pair classification, we generate predictions by applying an optimal threshold to the cosine similarity of the sentence pairs, measuring performance via Accuracy. Finally, for Semantic Textual Similarity (STS), we evaluate the alignment between the embeddings' cosine similarity and human-annotated gold standards using Spearman correlation. In our method, all models are fully fine-tuned using the unified objective $\mathcal{L}_{total}$. We evaluate the quality of Matryoshka representations across a fixed set of nested dimensions $\mathcal{D} = \{16, 32, 64, 128, 256, 512, 768/1024\}$. For Text Classification, we report F1 Score. For NLI tasks, we report Accuracy. For Semantic Textual Similarity (STS) benchmarks, we report the Spearman Correlation Coefficient.

\subsection{SIA and PIC hyperparameters}

\paragraph{Top-k Schedule Specification.}
To ensure reproducibility and robustness, we implement a deterministic linear schedule for the token selection threshold $k_i$. For our standard backbone configurations with a representation hierarchy $\mathcal{D} = \{16, 32, 64, 128, 256, 512, 768\}$, we align the first 6 lower-dimensional prefixes against the full-dimensional teacher. We set $k_i = \max(8, \lceil \gamma_i \cdot m \rceil)$, where $m$ is the sequence length. The ratio $\gamma_i$ increases monotonically with the dimension size: specifically, we utilize $\gamma = [0.2, 0.3, 0.4, 0.5, 0.6, 0.7]$ corresponding to the dimensions $[16, 32, 64, 128, 256, 512]$. This schedule ensures that the most compressed representations ($d=16$) focus on the top 20\% salient tokens, while larger prefixes progressively incorporate up to 70\% of the context, with a minimum floor of $k_{\min}=8$ tokens to preserve basic sentence structure in short inputs.

\paragraph{Layers and checkpoints applied in MIPIC}
In our framework, let $\mathcal{L}$ denote the set of layers applied in MIPIC, and $\mathcal{C}$ represent the set of checkpoints applied in PIC, defined as tuples $(d, l)$ where $d$ indicates the target dimension and $l$ corresponds to the layer index. The specific configurations of $\mathcal{L}$ and $\mathcal{C}$ are adapted to the architecture and depth of each model. For TinyBERT-6L, we utilize all layers, setting $\mathcal{L} = \{1, 2, 3, 4, 5, 6\}$ with checkpoints $\mathcal{C} = \{(16, 1),\allowbreak (32, 2),\allowbreak (64, 3),\allowbreak (256, 4),\allowbreak (512, 5),\allowbreak (768, 6)\}$. For BERT-base, we select distributed layers $\mathcal{L} = \{2, 4, 6, 8, 9, 10, 12\}$, associated with $\mathcal{C} = \{(16, 2),\allowbreak (32, 4),\allowbreak (64, 6),\allowbreak (128, 8),\allowbreak (256, 9),\allowbreak (512, 10),\allowbreak (768, 12)\}$. For the larger BGE-M3 model, we employ $\mathcal{L} = \{1, 4, 7, 11, 15, 19, 24\}$ and checkpoints $\mathcal{C} = \{(16, 1),\allowbreak (32, 4),\allowbreak (64, 7),\allowbreak (128, 11),\allowbreak (256, 15),\allowbreak (512, 19),\allowbreak (1024, 24)\}$. Finally, for Qwen3, the configuration is set to $\mathcal{L} = \{2, 6, 12, 16, 20, 24, 28\}$ with $\mathcal{C} = \{(16, 2),\allowbreak (32, 6),\allowbreak (64, 12),\allowbreak (128, 16),\allowbreak (256, 20),\allowbreak (512, 24),\allowbreak (1024, 28)\}$.

Our code is available at: \url{https://github.com/tmp0810/Matryoshkha-Representation-Learning-self-distilled}.

\section{Additional ablation study}
\label{more ablation}

\begin{table*}[htbp]
\centering
\caption{Performance comparison on in-domain datasets using BGEM3 and Qwen 3 (0.6B) backbones. \textbf{Bold} indicates the best performance for a given representation size.}
\label{tab:bgem3_qwen_results}
\resizebox{\textwidth}{!}{%
\begin{tabular}{ll ccc ccc c ll ccc ccc}
\toprule
 & & \multicolumn{3}{c}{\textbf{BGEM3}} & \multicolumn{3}{c}{\textbf{Qwen}} & & & & \multicolumn{3}{c}{\textbf{BGEM3}} & \multicolumn{3}{c}{\textbf{Qwen}} \\
\cmidrule(lr){3-5} \cmidrule(lr){6-8} \cmidrule(lr){12-14} \cmidrule(lr){15-17}
\textbf{Dataset} & \textbf{Dim} & MRL & ESE & MIPIC & MRL & ESE & MIPIC & & \textbf{Dataset} & \textbf{Dim} & MRL & ESE & MIPIC & MRL & ESE & MIPIC \\
\midrule

% Row 1: Banking77 (Left) & WiC (Right)
\multirow{7}{*}{Banking77} 
 & 16 & 73.06 & 72.07 & \textbf{75.38} & 55.02 & 56.94 & \textbf{58.45} & & \multirow{7}{*}{WiC} 
 & 16 & 61.35 & 61.53 & \textbf{61.92} & 61.71 & 62.92 & \textbf{63.71} \\
 & 32 & 86.09 & 84.91 & \textbf{86.48} & 77.72 & 78.29 & \textbf{78.51} & & 
 & 32 & 61.78 & 61.23 & \textbf{62.45} & 63.78 & 63.64 & \textbf{64.35} \\
 & 64 & 90.53 & 90.67 & \textbf{90.74} & 84.31 & 83.44 & \textbf{85.82} & & 
 & 64 & 62.14 & 63.12 & \textbf{63.91} & 64.57 & 64.35 & \textbf{66.28} \\
 & 128 & 92.10 & 92.02 & \textbf{92.94} & 87.87 & 87.25 & \textbf{88.84} & & 
 & 128 & 63.85 & 64.32 & \textbf{64.57} & 64.42 & 64.14 & \textbf{65.79} \\
 & 256 & 92.71 & 92.06 & \textbf{92.89} & 89.62 & 89.67 & \textbf{90.55} & & 
 & 256 & 64.21 & 64.28 & \textbf{64.69} & 65.00 & 64.07 & \textbf{66.21} \\
 & 512 & 92.93 & 92.13 & \textbf{92.98} & 91.61 & \textbf{91.42} & 91.32 & & 
 & 512 & 64.71 & 64.21 & \textbf{65.04} & 64.57 & 64.42 & \textbf{66.42} \\
 & 1024 & \textbf{93.11} & 92.25 & 92.81 & 91.81 & 91.98 & \textbf{92.16} & & 
 & 1024 & 65.38 & 64.42 & \textbf{65.58} & 63.28 & 64.14 & \textbf{66.07} \\
\cmidrule(r){1-8} \cmidrule(l){10-17}

% Row 2: TweetEval (Left) & SICK (Right)
\multirow{7}{*}{TweetEval} 
 & 16 & 56.91 & 55.32 & \textbf{58.08} & 55.15 & 53.35 & \textbf{57.96} & & \multirow{7}{*}{SICK} 
 & 16 & 69.66 & 68.35 & \textbf{69.89} & 64.15 & 64.45 & \textbf{66.37} \\
 & 32 & 61.72 & 59.95 & \textbf{61.82} & 61.83 & 59.11 & \textbf{62.13} & & 
 & 32 & 71.03 & 69.48 & \textbf{71.08} & 68.32 & 67.66 & \textbf{69.14} \\
 & 64 & 66.80 & 65.95 & \textbf{67.43} & 64.63 & 63.24 & \textbf{65.41} & & 
 & 64 & 71.63 & 70.27 & \textbf{71.92} & 69.12 & 68.92 & \textbf{69.89} \\
 & 128 & 70.06 & 69.67 & \textbf{70.98} & 67.51 & 65.99 & \textbf{68.35} & & 
 & 128 & 71.93 & 70.83 & \textbf{72.50} & 69.68 & 69.54 & \textbf{69.79} \\
 & 256 & 72.66 & 71.90 & \textbf{72.84} & 70.23 & 67.21 & \textbf{70.91} & & 
 & 256 & 72.46 & 71.08 & \textbf{72.84} & 69.45 & 69.74 & \textbf{69.82} \\
 & 512 & 72.61 & 72.69 & \textbf{72.98} & 71.09 & 69.67 & \textbf{71.83} & & 
 & 512 & 72.78 & 71.24 & \textbf{73.14} & 69.53 & 69.03 & \textbf{69.98} \\
 & 1024 & 72.95 & \textbf{74.13} & 73.56 & 71.82 & 69.32 & \textbf{71.94} & & 
 & 1024 & 72.77 & 71.23 & \textbf{73.15} & 69.33 & 69.09 & \textbf{69.76} \\
\cmidrule(r){1-8} \cmidrule(l){10-17}

% Row 3: MRPC (Left) & STSB (Right)
\multirow{7}{*}{MRPC} 
 & 16 & 67.21 & 67.43 & \textbf{67.65} & 68.11 & 67.18 & \textbf{69.69} & & \multirow{7}{*}{STSB} 
 & 16 & 75.21 & 73.96 & \textbf{76.81} & 67.83 & 67.66 & \textbf{68.44} \\
 & 32 & 67.36 & 67.65 & \textbf{67.69} & 68.34 & 67.24 & \textbf{69.15} & & 
 & 32 & 77.71 & 76.13 & \textbf{78.30} & 71.82 & 70.55 & \textbf{72.65} \\
 & 64 & 67.47 & 67.69 & \textbf{67.73} & 68.63 & 67.59 & \textbf{69.57} & & 
 & 64 & 79.50 & 77.61 & \textbf{80.30} & 73.22 & 72.39 & \textbf{74.36} \\
 & 128 & 67.39 & 67.47 & \textbf{67.71} & 69.04 & 67.53 & \textbf{69.10} & & 
 & 128 & 80.59 & 78.55 & \textbf{81.27} & 75.17 & 74.82 & \textbf{75.99} \\
 & 256 & 67.42 & \textbf{67.88} & 67.76 & 68.75 & 67.65 & \textbf{70.68} & & 
 & 256 & \textbf{82.44} & 79.40 & 82.20 & 75.67 & 74.05 & \textbf{75.92} \\
 & 512 & 67.71 & 67.59 & \textbf{67.89} & 70.45 & 67.88 & 70.21 & & 
 & 512 & 81.28 & 80.94 & \textbf{83.07} & 76.07 & 74.03 & \textbf{76.65} \\
 & 1024 & 67.59 & 67.53 & \textbf{67.97} & \textbf{70.98} & 68.05 & 70.73 & & 
 & 1024 & 82.52 & 80.52 & \textbf{83.39} & 75.76 & 74.29 & \textbf{76.85} \\
\bottomrule
\end{tabular}%
}
\end{table*}

\begin{table*}[htbp]
\centering
\caption{Performance comparison on out-of-domain datasets using BGEM3 and Qwen 3 (0.6B) backbones. \textbf{Bold} indicates the best performance for a given representation size.}
\label{tab:out_domain_results}
\resizebox{\textwidth}{!}{%
\begin{tabular}{ll ccc ccc c ll ccc ccc}
\toprule
 & & \multicolumn{3}{c}{\textbf{BGEM3}} & \multicolumn{3}{c}{\textbf{Qwen}} & & & & \multicolumn{3}{c}{\textbf{BGEM3}} & \multicolumn{3}{c}{\textbf{Qwen}} \\
\cmidrule(lr){3-5} \cmidrule(lr){6-8} \cmidrule(lr){12-14} \cmidrule(lr){15-17}
\textbf{Dataset} & \textbf{Dim} & MRL & ESE & MIPIC & MRL & ESE & MIPIC & & \textbf{Dataset} & \textbf{Dim} & MRL & ESE & MIPIC & MRL & ESE & MIPIC \\
\midrule

% Row 1: STS12 (Left) & STS16 (Right)
\multirow{7}{*}{STS12} 
 & 16 & 62.95 & 61.97 & \textbf{63.87} & 59.41 & 58.10 & \textbf{62.96} & & \multirow{7}{*}{STS16} 
 & 16 & 73.16 & 72.77 & \textbf{74.56} & 67.39 & 67.98 & \textbf{68.17} \\
 & 32 & 67.02 & 64.78 & \textbf{67.67} & 63.70 & 61.15 & \textbf{65.71} & & 
 & 32 & 74.67 & 75.54 & \textbf{75.68} & 70.88 & 70.12 & \textbf{71.85} \\
 & 64 & 67.99 & 65.85 & \textbf{68.51} & 65.93 & 63.30 & \textbf{66.61} & & 
 & 64 & 76.96 & 77.10 & \textbf{78.13} & 72.84 & 71.63 & \textbf{74.41} \\
 & 128 & 69.38 & 66.83 & \textbf{69.90} & 67.31 & 66.89 & \textbf{68.28} & & 
 & 128 & 78.88 & 77.75 & \textbf{79.26} & 74.29 & 73.28 & \textbf{75.85} \\
 & 256 & 70.82 & 67.73 & \textbf{71.06} & 68.48 & 69.68 & \textbf{70.93} & & 
 & 256 & 80.91 & 78.48 & \textbf{80.93} & 75.00 & 74.71 & \textbf{76.67} \\
 & 512 & 71.64 & 68.80 & \textbf{71.76} & 68.93 & 68.01 & \textbf{70.85} & & 
 & 512 & 82.13 & 80.20 & \textbf{82.21} & 75.26 & 75.61 & \textbf{77.31} \\
 & 1024 & 72.11 & 68.56 & \textbf{72.45} & \textbf{70.74} & 69.28 & 70.31 & & 
 & 1024 & \textbf{82.31} & 79.97 & 82.10 & 74.60 & 75.80 & \textbf{76.83} \\
\cmidrule(r){1-8} \cmidrule(l){10-17}

% Row 2: STS13 (Left) & SickR (Right)
\multirow{7}{*}{STS13} 
 & 16 & 75.32 & 73.32 & \textbf{76.04} & 67.83 & 68.09 & \textbf{69.23} & & \multirow{7}{*}{SickR} 
 & 16 & 70.49 & 69.87 & \textbf{72.42} & 63.81 & 63.52 & \textbf{65.62} \\
 & 32 & 77.20 & 75.60 & \textbf{78.76} & 70.71 & 70.58 & \textbf{73.59} & & 
 & 32 & 73.43 & 72.06 & \textbf{74.83} & 67.81 & 66.95 & \textbf{68.32} \\
 & 64 & 79.66 & 77.88 & \textbf{80.02} & 73.65 & 73.23 & \textbf{76.94} & & 
 & 64 & 75.43 & 72.76 & \textbf{76.64} & 69.08 & 68.96 & \textbf{69.69} \\
 & 128 & 80.96 & 78.98 & \textbf{81.15} & 75.27 & 75.89 & \textbf{77.35} & & 
 & 128 & 76.52 & 75.12 & \textbf{77.31} & 69.75 & 69.63 & \textbf{69.87} \\
 & 256 & 82.42 & 80.30 & \textbf{82.49} & 77.09 & 76.16 & \textbf{77.48} & & 
 & 256 & 77.03 & 76.41 & \textbf{77.34} & 69.63 & 69.89 & \textbf{69.96} \\
 & 512 & 83.54 & 81.76 & \textbf{83.68} & 78.52 & 77.82 & \textbf{78.63} & & 
 & 512 & \textbf{77.56} & 76.55 & 77.43 & \textbf{69.76} & 69.14 & 69.75 \\
 & 1024 & 84.03 & 81.79 & \textbf{84.97} & \textbf{79.51} & 78.51 & 78.82 & & 
 & 1024 & \textbf{77.68} & 76.48 & 77.45 & 69.56 & \textbf{69.65} & 69.53 \\
\cmidrule(r){1-8} \cmidrule(l){10-17}

% Row 3: STS14 (Left) & Emotion (Right)
\multirow{7}{*}{STS14} 
 & 16 & 67.07 & 65.03 & \textbf{68.33} & 59.40 & 59.97 & \textbf{60.85} & & \multirow{7}{*}{Emotion} 
 & 16 & 29.92 & 28.46 & \textbf{30.83} & 29.39 & 28.62 & \textbf{31.35} \\
 & 32 & 69.43 & 67.28 & \textbf{70.65} & 61.20 & 62.14 & \textbf{63.14} & & 
 & 32 & 36.94 & 33.94 & \textbf{38.94} & 39.93 & 39.61 & \textbf{41.92} \\
 & 64 & 72.46 & 68.96 & \textbf{72.73} & 64.43 & 63.68 & \textbf{65.43} & & 
 & 64 & 43.45 & 44.53 & \textbf{46.08} & 47.44 & \textbf{48.42} & 48.03 \\
 & 128 & 73.38 & 70.18 & \textbf{73.83} & 66.01 & 66.95 & \textbf{67.76} & & 
 & 128 & 49.84 & 50.05 & \textbf{52.40} & 54.75 & 51.51 & \textbf{55.61} \\
 & 256 & 74.45 & 71.39 & \textbf{74.75} & 67.44 & 67.32 & \textbf{68.94} & & 
 & 256 & 54.07 & 53.41 & \textbf{57.33} & 58.44 & \textbf{59.33} & 58.77 \\
 & 512 & \textbf{74.93} & 72.45 & 72.28 & \textbf{68.73} & 68.57 & 68.06 & & 
 & 512 & 58.73 & 55.59 & \textbf{60.67} & 62.81 & \textbf{64.79} & 61.52 \\
 & 1024 & 75.33 & 72.35 & \textbf{75.65} & 69.59 & 69.14 & \textbf{69.97} & & 
 & 1024 & 61.39 & 60.12 & \textbf{62.45} & \textbf{66.42} & 66.26 & 65.15 \\
\cmidrule(r){1-8} \cmidrule(l){10-17}

% Row 4: STS15 (Left) & SciTail (Right)
\multirow{7}{*}{STS15} 
 & 16 & 74.57 & 75.44 & \textbf{76.79} & 67.39 & 68.85 & \textbf{71.61} & & \multirow{7}{*}{SciTail} 
 & 16 & 80.62 & 80.47 & \textbf{82.47} & 80.04 & 80.49 & \textbf{82.42} \\
 & 32 & 78.46 & 77.98 & \textbf{79.05} & 70.88 & 71.11 & \textbf{73.85} & & 
 & 32 & 81.74 & 81.32 & \textbf{84.24} & 82.59 & 81.74 & \textbf{83.44} \\
 & 64 & 81.39 & 80.14 & \textbf{81.94} & 72.84 & 72.42 & \textbf{75.14} & & 
 & 64 & 83.58 & 81.60 & \textbf{84.51} & 83.67 & 82.32 & \textbf{84.23} \\
 & 128 & 82.29 & 80.46 & \textbf{83.10} & 74.29 & 74.29 & \textbf{77.66} & & 
 & 128 & 84.43 & 82.03 & \textbf{84.76} & 84.24 & 83.50 & \textbf{84.69} \\
 & 256 & 82.90 & 81.10 & \textbf{83.74} & 75.00 & 75.25 & \textbf{78.83} & & 
 & 256 & 84.19 & 82.03 & \textbf{84.57} & 84.36 & 83.55 & \textbf{84.79} \\
 & 512 & 83.19 & 81.45 & \textbf{84.22} & 75.26 & 76.44 & \textbf{79.04} & & 
 & 512 & 85.03 & 82.30 & \textbf{85.08} & 84.38 & 84.16 & \textbf{84.65} \\
 & 1024 & 83.62 & 81.69 & \textbf{84.45} & 74.60 & 76.86 & \textbf{79.12} & & 
 & 1024 & \textbf{85.46} & 82.69 & 85.13 & 84.28 & 84.21 & \textbf{84.54} \\
\bottomrule
\end{tabular}%
}
\end{table*}

The expansion of our experiments to high-capacity models, specifically Qwen3-0.6B embedding and BGE-M3, confirms the scalability and robustness of the MIPIC framework. The results demonstrate that the synergy between Self-Distilled Intra-Relational Alignment (SIA) and Progressive Information Chaining (PIC) remains highly effective as model parameters and depth increase, transitioning successfully from standard encoders to large-scale architectures. Extensive evaluations across nearly all benchmarks indicate that MIPIC consistently outperforms established baselines such as MRL and ESE. This persistent superiority suggests that our global chaining strategy effectively captures and organizes the complex semantic structures inherent in larger models, ensuring that task-specific signals are preserved throughout the network's internal representation hierarchy regardless of the model size. Key observations from the large-scale evaluation highlight the remarkable resilience of MIPIC under extreme dimensional compression and its enhanced generalization capabilities. The performance gap between MIPIC and other methods is most pronounced at extremely low dimensions, such as 16 and 32, where it effectively condenses vast knowledge into compact prefixes while maintaining high task awareness. Furthermore, on out-of-domain (OOD) datasets like STS12--16 and SciTail, MIPIC maintains a consistent accuracy lead over current state-of-the-art baselines. This confirms that the depth-wise semantic consolidation provided by PIC allows large backbones to retain a stable semantic bridge, ensuring that even the most compressed representations remain grounded in robust, generalized knowledge across diverse tasks and domains.

\section{Use of Large Language Models}

We acknowledge the use of Large Language Models (LLMs) during the preparation of this manuscript. These tools were utilized strictly for stylistic refinement, grammatical editing, and improving the overall flow of the presentation. At no point were LLMs used to conceptualize research ideas, formulate hypotheses, or interpret experimental findings. The authors remain fully accountable for the integrity and accuracy of the final content.

\end{document}